\documentclass[lettersize,journal]{IEEEtran}
\usepackage{amsmath,amsfonts}
\usepackage{algorithmic}
\usepackage{algorithm}
\usepackage{array}
\usepackage[caption=false,font=normalsize,labelfont=sf,textfont=sf]{subfig}
\usepackage{textcomp}
\usepackage{stfloats}
\usepackage{url}
\usepackage{verbatim}
\usepackage{graphicx}
\usepackage{cite}
\usepackage{gensymb}
\begin{document}

\title{Investigating the Effect of a Series Elastic Actuation Retrofit to Black-Box Actuators}

\author{I. Tregear, A. Aktas, F. Rodriguez y Baena 

\thanks{I. Tregear was with Imperial College London, Mechanical Engineering Department, UK. He is now with KAIKAKU.AI, London, N1 6NG, UK. Email: \texttt{ivantregear@gmail.com}}

\thanks{A. Aktas and F. Rodriguez y Baena are with Imperial College London, Mechanical Engineering Department, SW7 2AZ, UK. The work of Ayhan Aktas was supported by the Republic of T{\"u}rkiye.}
}


\markboth{Preprint}%
{Author \MakeLowercase{\textit{et al.}}: Paper Title}

\maketitle

\begin{abstract}
In robotic applications, actuators are typically designed to be stiff with minimal backlash to ensure precision and repeatability. However, this limits compliance, leading to potential damage and poor force control in uncertain environments. Series Elastic Actuation (SEA) introduces compliance to enhance disturbance rejection and enable force measurement via Hooke’s Law but reduces system bandwidth.

A custom Series Elastic (SE) element was retrofitted to a black-box actuator to mitigate non-linearities like backlash and static friction. Integrating the SE element enabled high-fidelity force measurements, improving force control bandwidth and performance.

A torsional SE element was designed through Finite Element (FE) analysis, yielding a stiffness of \(2155.4\) Nm/rad. Open-loop force control bandwidth was measured for the original motor and the SEA-integrated configuration, while closed-loop bandwidth was assessed using feedback from the SEA and a commercial force sensor. The SEA module increased bandwidth from \(10.32\) Hz to \(30.32\) Hz, a \(2.93\times\) improvement. Additionally, it outperformed the commercial sensor by \(7.63\%\) despite costing 25 GBP, a fraction of the price.

\end{abstract}


\section{Introduction}
\IEEEPARstart{S}{eries} Elastic Actuators (SEAs) introduce compliance into actuators to enhance disturbance rejection and enable force measurement via Hooke’s Law \cite{Williamson1995SeriesActuators, Robinson1999SeriesRobot, Sergi2012DesignWalking}. However, this reduces system bandwidth due to a lower natural frequency \cite{Williamson1995SeriesActuators}. As a result, SEAs are used in applications where rejecting high-frequency impacts is critical, such as rehabilitative exoskeletons \cite{Rampeltshammer2020AnActuators, Lagoda2010DesignTraining, Muller2015ModellingOrthoses, DosSantos2015Design}.

Research has focused on SEA control architectures, particularly impedance control, due to their lower impedance compared to rigid actuators \cite{Zhao2018ImpedanceActuators, Tagliamonte2010DesignApplications, Pratt2004LateActuators}. Other strategies include velocity-sourced control \cite{Wyeth2006ControlActuators}, acceleration-based impedance control \cite{Calanca2017ImpedanceControl}, and disturbance observers \cite{Rampeltshammer2020AnActuators}. SEA performance is also influenced by mechanical design \cite{Townsend1989MechanicalDesign}, with SE element placement affecting force sensitivity, compliance, and transmissibility \cite{Paine2014DesignActuators}. Lee et al. generalized SEA configurations into Force Sensing SEA (FSEA), Reaction FSEA (RFSEA), and Transmitted FSEA (TFSEA) \cite{Lee2017GeneralizationComparison}:

\begin{itemize}
  \item FSEA: spring placed after gear-train.
  \item RFSEA: spring placed between motor and mechanical ground or between motor and gear-train.
  \item TFSEA: spring placed within gear-train. 
\end{itemize}

TFSEAs are rarely used due to integration complexity \cite{Yun2020}. RFSEAs, more common \cite{Kim2019}, allow direct deflection measurement from a single datum and simplify integration but increase inertia, reducing force-sensing sensitivity \cite{Park2016}. FSEAs, though mechanically complex, improve force sensitivity by eliminating inertia-induced measurement errors \cite{Paine2014DesignActuators}.

While SE placement effects on bandwidth remain underexplored, most studies focus on bandwidth as a function of force or torque amplitude \cite{Leal2016}. Additionally, direct SE element deflection measurement improves accuracy over differential methods, which are more affected by actuator non-linearities like gear-train backlash \cite{Williamson1995SeriesActuators, Robinson2000DesignControl}.

SE element design is crucial to actuator performance \cite{Baldoni2018}. Laser-cut torsional springs, first proposed by Sergi et al. \cite{Sergi2012DesignWalking}, are now widely used in rotary applications \cite{Irmscher2018DesignActuators, Carpino2012ARobots, Lagoda2010DesignTraining, DosSantos2015Design} due to their cost-effectiveness and customizability compared to traditionally manufactured springs \cite{Valsange2012}. These elements enable precise geometries, rapid prototyping, and fine-tuning of torque capacity, stiffness, and linear range.

A key application of SEAs is biomimetic robots, such as quadrupeds, which use quasi-direct drive actuators with low reduction ratios (\(<10\)) for improved responsiveness by reducing reflected inertia \cite{Katz2018ARobots, Singh2020, Gealy2019}. Compliance integration enhances performance in unstructured environments, as demonstrated in SEA-driven robot dog legs, where torque amplitude was linked to SEA bandwidth, achieving \(32 - 35\) Hz \cite{Lee2019DevelopmentLeg}.

SEAs are typically designed with early-stage integration, optimizing stiffness, compliance, torque capacity, and efficiency. However, many real-world applications involve pre-existing actuators that require retrofitting, a challenge largely overlooked in the literature—particularly regarding non-linearities such as static friction and backlash.

This study retrofits a custom SEA onto an all-in-one motor actuator, mitigating key non-linearities and enhancing closed-loop force control bandwidth. A planar torsional disc framework was developed and validated through simulation and experimentation. The study identifies optimal retrofitted SEA configurations, emphasizing FSEA designs and direct displacement measurement. By addressing these challenges, this work provides insights into retrofittable SEA designs and explores potential bandwidth improvements in actuators not originally designed for SEA integration. Section \ref{Methods and Materials} details the methods and experimental setup, Section \ref{Results} presents results, and Sections \ref{Discussion} and \ref{Conclusion} discuss findings and future work.

\section{Methods and Materials} \label{Methods and Materials}
\subsection{Design Requirements} \label{Design Requirements}

This study evaluates the impact of retrofitting an SE element on an actuator’s closed-loop force control bandwidth. An all-in-one actuator (motor, driver, encoder, transmission) was chosen as a baseline to isolate the SE element’s effect, distinguishing this work from prior SEA implementations that integrate SE elements during initial design.

The RMD X8-V2 (MyActuator, China) \cite{MyActuator2021MyDownloads}, a derivative of the MIT-Cheetah actuator, was selected. Table \ref{tab:rmd_x8_specifications} lists its nominal specifications, which were re-measured to account for manufacturing variability. Notably, backlash was found to be \(4 \times\) higher than advertised, likely due to vibration-induced fretting on gear teeth during shipment \cite{Biii1982}. While inductance and resistance were not directly measurable due to inaccessible coils, the electrical time constant was measured to infer their ratio.

\begin{table}[h!]
  \centering
  \caption{RMD X8 V2 Specifications \cite{MyActuatorRMDManual}}
  \label{tab:rmd_x8_specifications}
  \begin{tabular}{cccc}
    Specification & Provided Value & Measured Value\\
    \hline
    Reduction Ratio & \(9:1\) & -\\
    Nominal Torque [Nm] & \(9\) & -\\
    Phase to Phase Inductance [mH] & \(0.22\) & -\\
    Resistance [$\Omega$] & \(0.4\) & -\\
    Electrical Time Constant & $5.5\times10^{-4}$ & $0.42\times10^{-4}$\\
    Torque Constant [NmA\textsuperscript{-1}] & \(2.09\) & \(1.393\)\\
    Rotor Inertia [kgm\textsuperscript{2}] & \(0.00026\) & -\\
    Backlash [$\degree$] & \(0.083\) & \(0.3336\)\\
  \end{tabular}
\end{table}

Key design principles were drawn from Robinson's thesis \cite{Robinson2000DesignControl} on SEAs for closed-loop force control. The SE element was selected, its measurement method defined, and its placement determined. Direct angular measurement was prioritized for accuracy using an AS5311 Integrated Circuit (IC) (AMS OSRAM, Austria) with \(2\) mm pitch magnetic encoder tape, achieving \(1.109\times10^{-5}\) rad/pulse resolution. The encoder was mounted on a custom PCB attached to the motor actuator output shaft, rotating with both the motor and SEA output.

\begin{figure}[h!]
    \centering
    \includegraphics[width=\columnwidth]{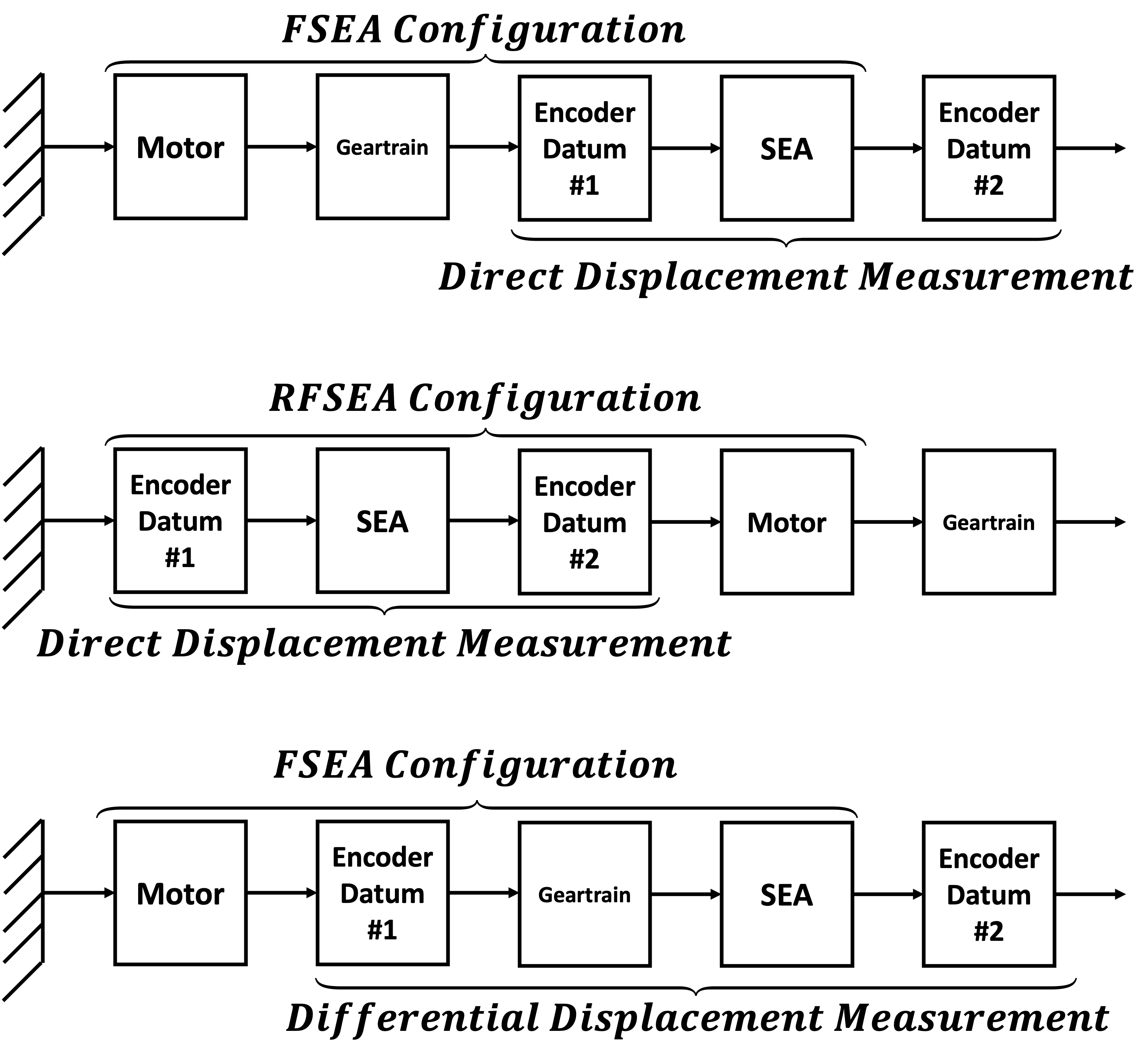}
    \caption{Block diagram comparison of different SEA configurations. FSEA with direct displacement measurement (top), RFSEA with direct displacement measurement (centre), FSEA with differential displacement measurement (bottom).}
    \label{fig:sea_configuration_comparison_block_diagram_large_text}
\end{figure}

TFSEA was ruled out due to the actuator’s integrated gearbox. While RFSEA simplifies mechanical implementation by fixing the SE element to mechanical ground \cite{Paine2014DesignActuators}, FSEA was chosen for its superior force sensitivity, defined as the ratio of SE element deflection to external force \cite{Lee2017GeneralizationComparison}. The selected FSEA with direct displacement measurement is shown in Figure \ref{fig:sea_configuration_comparison_block_diagram_large_text}, along with comparisons to RFSEA and FSEA with differential displacement measurement.

\begin{figure}[h!]
    \centering
    \includegraphics[width=\columnwidth]{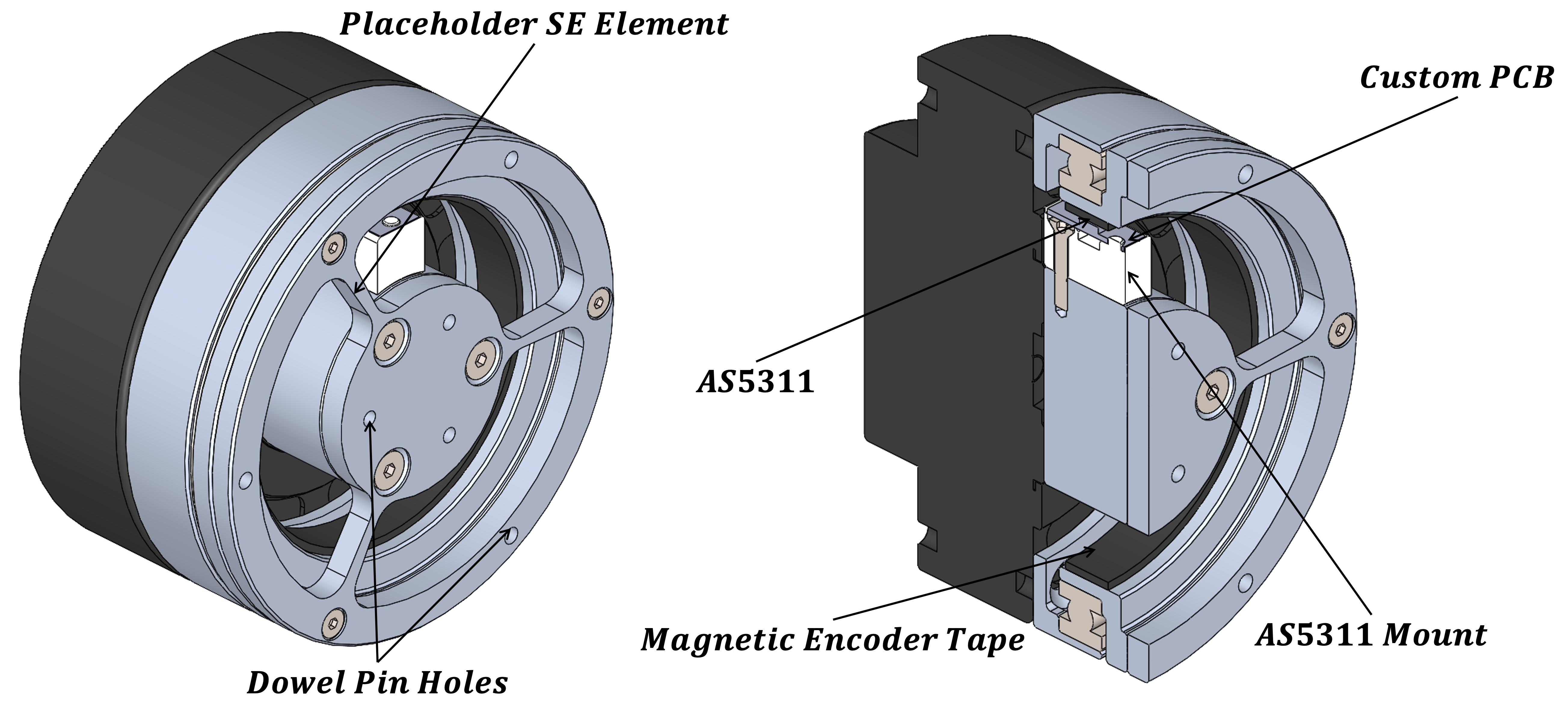}
    \caption{Final SEA proof of concept design with generic three-spoke SE element (left) and section view showing mounting for AS5311 magnetic encoder (right). A custom PCB positions the AS5311 below the magnetic tape, and is mounted directly to the motor output shaft, allowing it to rotate with the motor output and SEA.}
    \label{fig:final_se_design}
\end{figure}

Geometric relationships determined the encoder pulse count to position, but calibration corrected tape diameter variations. A calibration coefficient was obtained by averaging pulse counts over 15 motor rotations of \(2 \degree\) in both directions. A similar method measured actuator backlash by locking the motor with holding torque and manually moving the output while recording pulse changes. Backlash was \(0.334\degree\), approximately four times the manufacturer’s stated \(0.0833\degree\), likely due to vibration-induced fretting of the gear teeth \cite{Biii1982}. However, as backlash was not used in later analyses, this discrepancy did not affect the study. The final design is shown in Figure \ref{fig:final_se_design}.

\subsection{Series Elastic Element Design} \label{Series Elastic Element Design}

A laser-cut torsional disc was selected as the SE element, based on an established design framework \cite{Sergi2012DesignWalking, Lagoda2010DesignTraining, Carpino2012ARobots, DosSantos2015Design}. 

\begin{figure}[h!]
    \centering
    \includegraphics[width=\columnwidth]{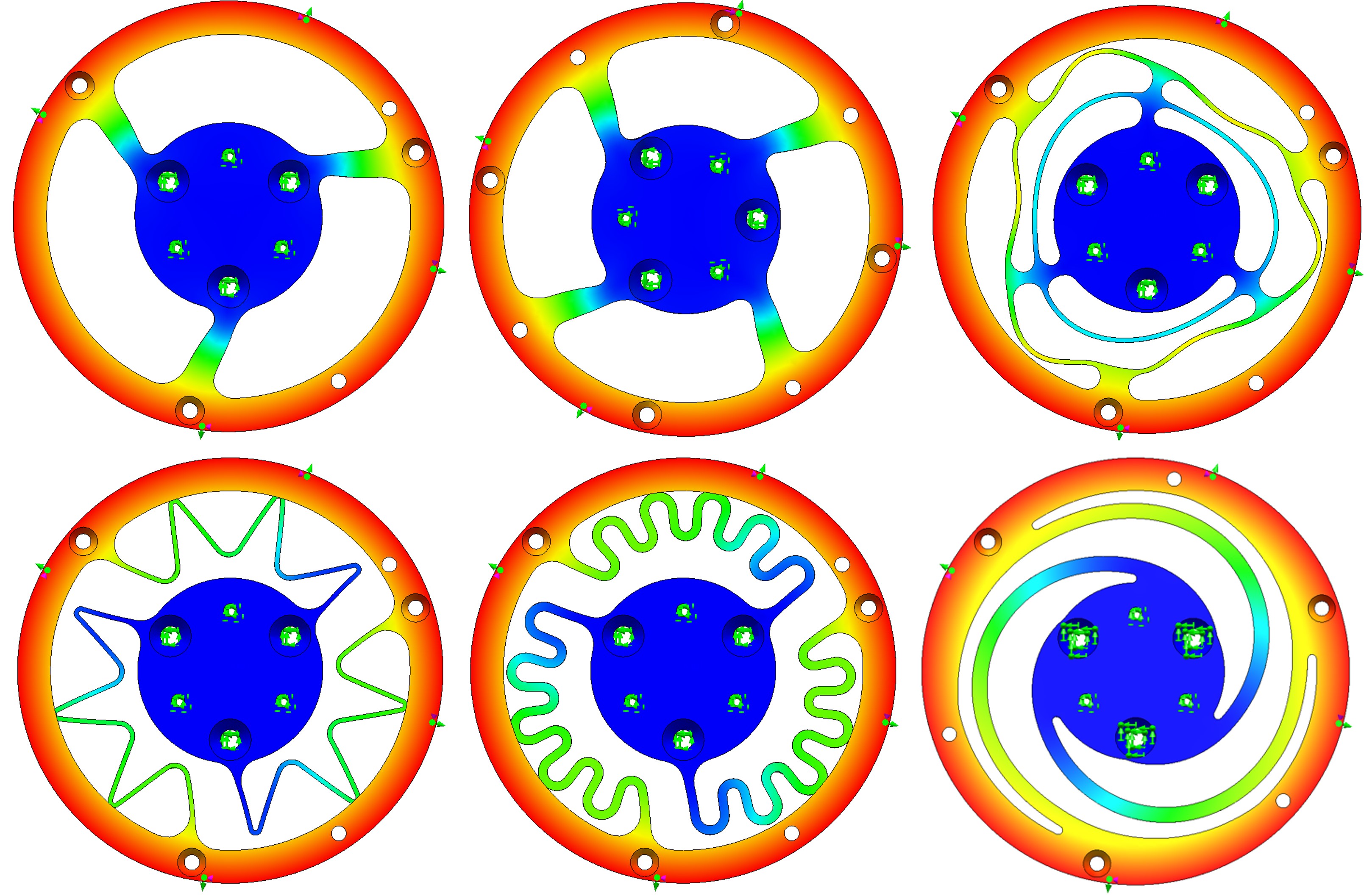}
    \caption{Chronological topology iterations from left to right, shown in the deformed state under 1 Nm load.}
    \label{fig:topology_iterations}
\end{figure}

A linear plane-strain study in SolidWorks (Dassault Systèmes, France) was employed to analyze a disc constrained at its center, with a tangential force applied at the circumference to simulate torque. Validation was performed by comparing results to an analytical in-plane torsion model \cite{Cwiekala2022}. This study identified a spring topology ensuring linear displacement within the motor torque range, as shown in Figure \ref{fig:topology_iterations}.

A subsequent non-linear plane-stress study was used to refine stiffness prediction and geometry. A triangular mesh with local refinement around mounting holes and stress concentrations was generated, with node count validated via a convergence study.

\subsection{Experimental Design} \label{Experimental Design}

\begin{figure}[h!]
    \centering
    \includegraphics[width=\columnwidth]{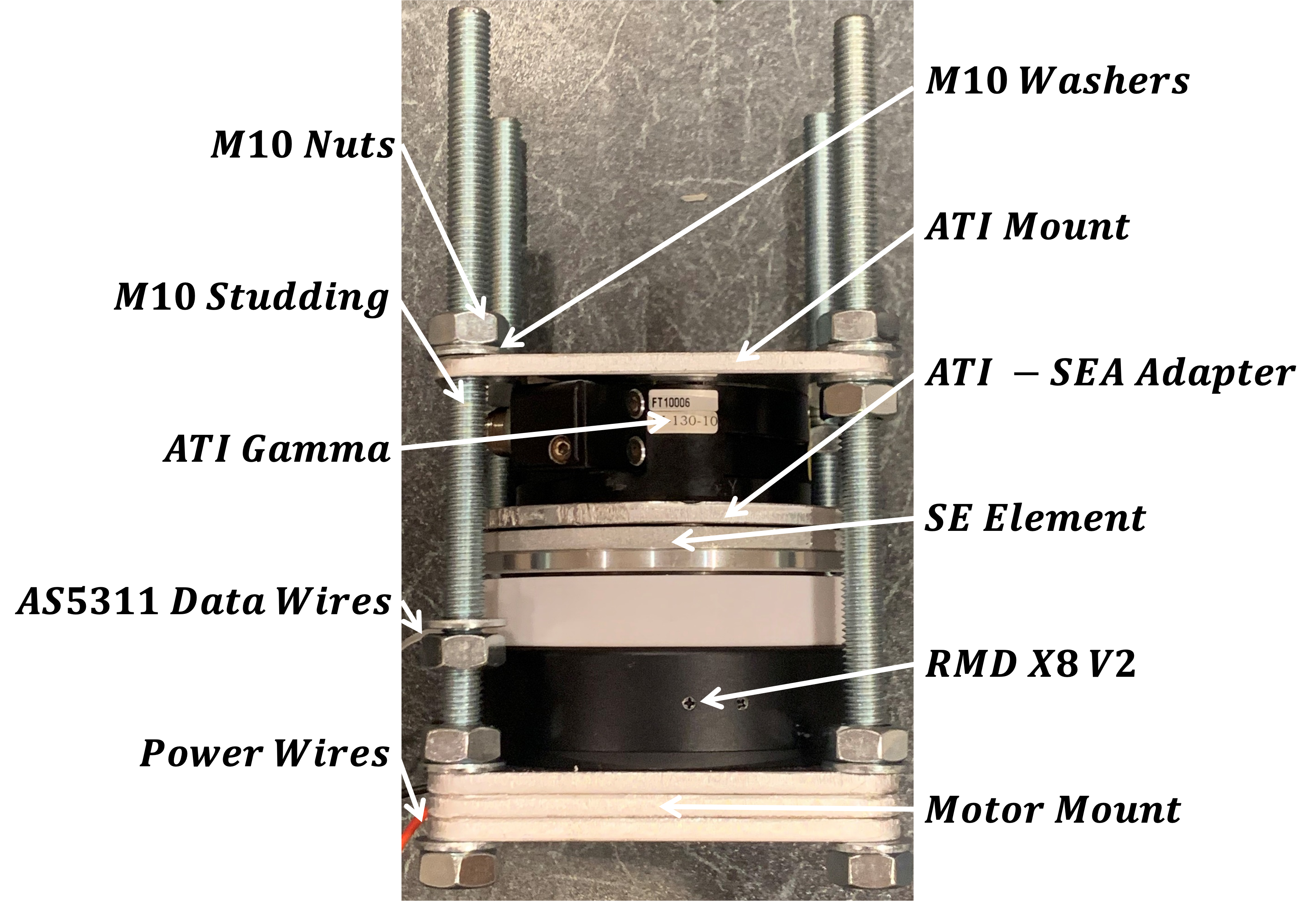}
    \caption{Setup for fixed load torque testing. The motor housing is rigidly attached to one sheet metal mount, and the SEA output is rigidly attached to another. Rotation between these is constrained via four lengths of M10 studding. In this way, the motor torque is entirely used to deform the SE element, isolating the comparison between applied motor torque and measured SE element torque.}
    \label{fig:test_setup}
\end{figure}

A Bode plot was produced to assess phase shift and force gain using an ATI Gamma force sensor (ATI Industrial Automation, USA), detailed in Table \ref{tab:ati_gamma_specifications}. To isolate inertia effects, the motor was tested in a fixed-load setup (Figure \ref{fig:test_setup}) while powered at 12V, current limited to 6A.

\begin{table}[h!]
  \centering
  \caption{ATI Gamma specifications \cite{ATIGamma}}
  \label{tab:ati_gamma_specifications}
  \begin{tabular}{c c}
    Specification & Value\\
    \hline
    \(T_z\) Range & $\pm{10}$ Nm \\
    \(T_{z, max}\) & $\pm{82}$ Nm \\
    \(T_z\) Resolution & \(125\) mNm \\
    \(k_{\theta}\) Stiffness & \(16,400\) Nm/rad \\
    \(f_{s}\) Sampling Rate & \(1400\) Hz \\
  \end{tabular}
\end{table}

The motor was tested in four configurations, subject to a linear frequency sweep, Equation \ref{eq:12}, at varying torque amplitudes.

\begin{enumerate}
    \item Original motor (open loop motor bandwidth)
    \item Motor with SEA Module, without feedback from SEA (open loop SEA bandwidth)
    \item Motor with SEA Module, \textit{with} SEA feedback (closed loop SEA bandwidth)
    \item Motor with SEA Module, with feedback from force sensor 
\end{enumerate}

\begin{equation} 
    \label{eq:12}
    T(t)=T_{max}\sin\left(2\pi\left(f_0t + \frac{f_1-f_0}{2T}t^2\right)\right)
\end{equation}

Bandwidth is traditionally defined as the frequency where the Bode plot magnitude response first drops below -3dB \cite{Mojahed2022}. This study considers a broader frequency range where effective control is maintained. The SEA introduces a second-order dynamic response, adding a slow pole due to the mass-spring system, sometimes causing a magnitude dip before recovery.

In such cases, a brief drop below -3dB does not necessarily degrade system performance at higher frequencies. To address this, a modified bandwidth definition is proposed, detailed in Table \ref{tab:bandwidth_definition}. While unchanged for continuously decreasing responses, additional leniency is applied to plots with temporary dips, better reflecting practical operability.

\begin{table}[h!]
  \centering
  \caption{Modified algorithm for bandwidth estimation.}
  \begin{tabular}{p{0.01\linewidth}p{0.45\linewidth}p{0.45\linewidth}}
    \hline
    1 & Identify DC gain & \(\left|H_{DC}\right|=\left|H\right|_{\omega \to 0}\)\\
    2 & Identify all frequencies where magnitude is greater than \(\left|H_{DC}\right| - 3\) dB & 
    \(\omega_{-3dB} = \{\omega : \omega\left(\left|H\right| \geq \left|H_{DC}\right| -3 \text{ dB} \right)\}\) \\
    2 & Identify frequency at global phase minimum & \(\omega\left(\phi_{min}\right) = \omega\left(\min\left(\phi\right)\right)\) \\
    3 & Split phase into those occurring before and after \(\phi_{min}\) & \(\phi_1=\{\phi : \phi \leq \phi_{min}\}\) \newline \(\phi_2=\{\phi : \phi \geq \phi_{min}\}\) \\
    4 & Find maximum phase in \(\phi_1\) & \(\phi_{1, max} = \max\left(\phi_1\right)\) \\
    5 & Find lowest frequency where \(\phi_2\) first surpasses \(\phi_{1, max}\) & \(\omega_{c} = \min\left(\omega\left(\phi_2 \geq phi_{1, max} \right)\right)\) \\
    6 & Select highest frequency below \(\omega_{c}\) where magnitude is greater than \(-3\) dB as the bandwidth & \(\omega_B = \max\left(\{\omega : \omega_{-3\text{ dB}} \leq \omega_c\}\right)\) \\
  \end{tabular}
  \label{tab:bandwidth_definition}
\end{table}

\subsection{Control} \label{Control}

\begin{figure}[h!]
    \centering
    \includegraphics[width=\columnwidth]{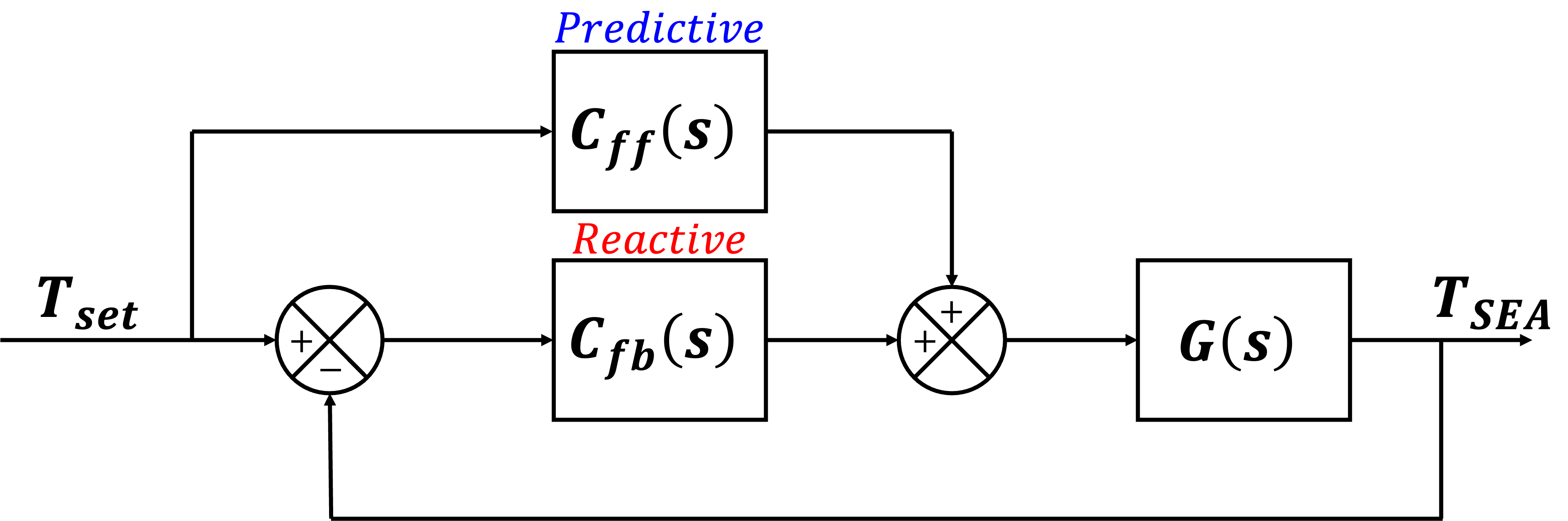}
    \caption{Block diagram of a combined feedforward-feedback controller architecture.}
    \label{fig:feedforward_block_diagram}
\end{figure}

A combined feedforward-feedback control architecture was implemented to close the force control loop, leveraging the fast response of feedforward control and the error correction of feedback. The block diagram is shown in Figure \ref{fig:feedforward_block_diagram}. For simplicity, both controllers were set to unity gain ($C_{ff}(s) = 1$ and $C_{fb}(s) = 1$).

Implemented in Python, the controller communicated with an ATmega 328p over serial bus, which interfaced with the motor actuator via CAN. To approximate continuous-time behaviour, the controller frequency was set above a minimum threshold to mitigate discrete-time delays. Following Franklin et al. \cite{Franklin2010FeedbackEdition}, the sampling frequency was required to be \(>20 \times\) the slowest system time constant. System identification of the step response to a force input measured this as \(0.04178\), leading to a control frequency threshold of \(479\) Hz.

\section{Results} \label{Results}

\begin{figure}[h!]
    \centering
    \includegraphics[width=\columnwidth]{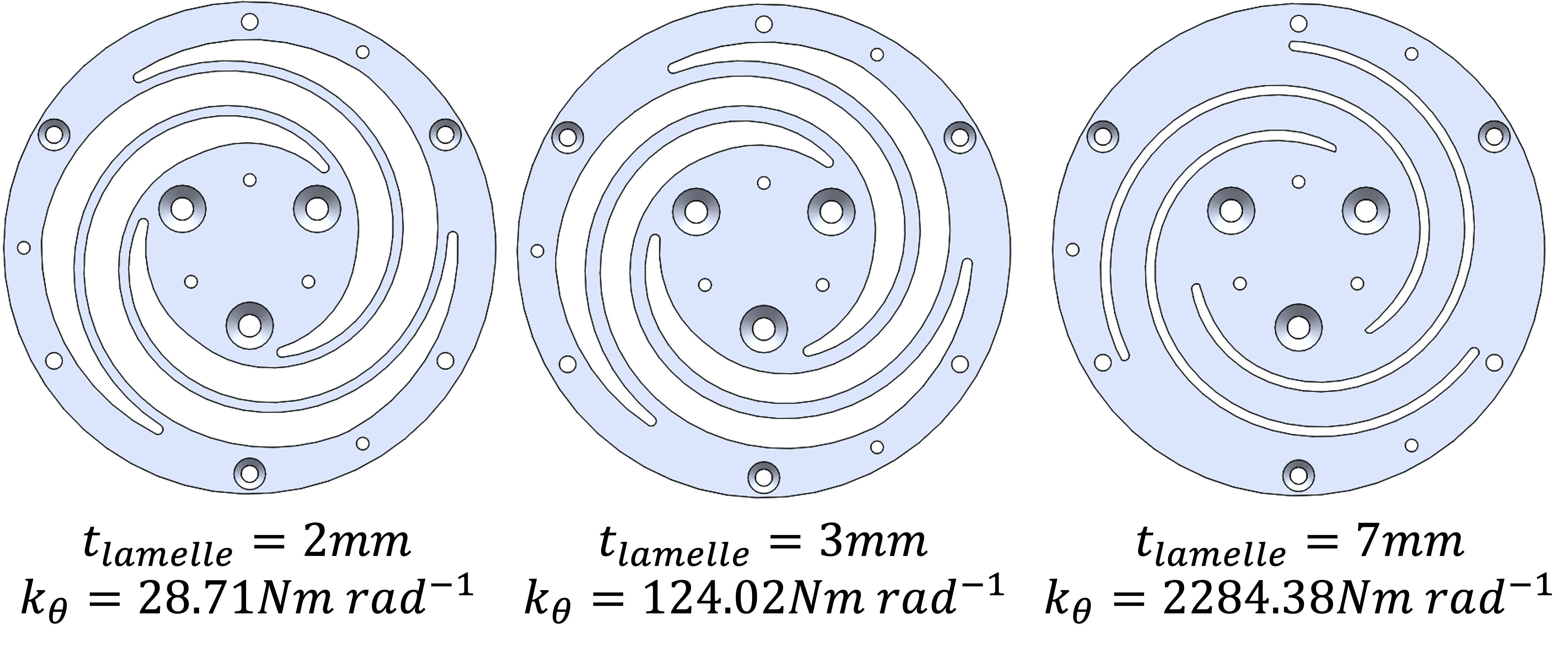}
    \caption{Final SE designs with lamella thicknesses of \(2\) mm (left), \(3\) mm (centre) and \(7\) mm (right), along with predicted stiffness.}
    \label{fig:final_se_topologies}
\end{figure}

The SE element topology was selected using a two-part FE framework. The final design was an order-3 axisymmetrical disc with spiral-shaped lamellar structures. Torsional rigidity was optimized by varying lamella thickness, with three candidates shown in Figure \ref{fig:final_se_topologies}. The \(7\) mm lamella disc was chosen for its linear deformation across the force sensor’s torque range. The manufactured SE element's torsional rigidity was \(2155.4\) Nm/rad, deviating \(-5.65\)\% from FE predictions. Controller frequency reached \(603\) Hz, exceeding the \(479\) Hz requirement by \(25.8\)\%.

\begin{figure*}[h!]
    \centering
    \includegraphics[width=\textwidth]{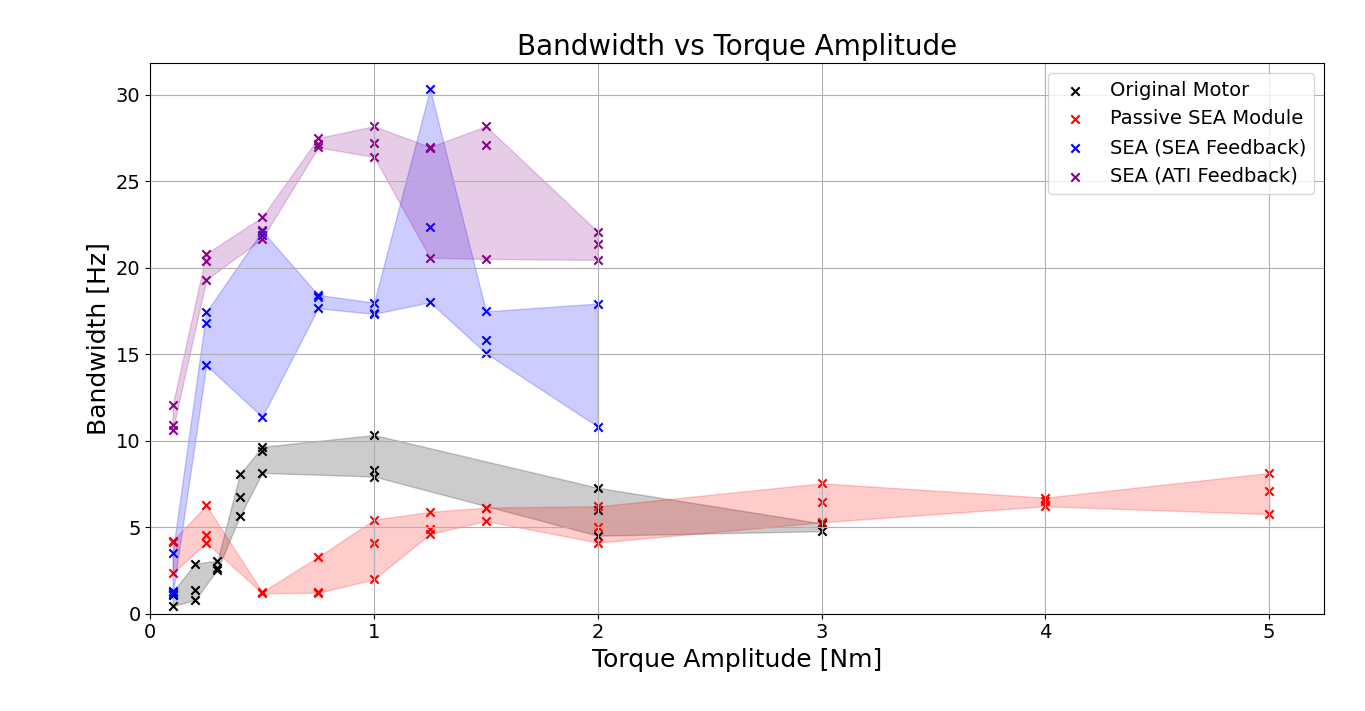}
    \caption{Bandwidth versus torque amplitude for four tested configurations: Original Motor, Passive SEA Module, SEA (SEA Feedback), and SEA (force sensor feedback). Shaded regions represent variability across trials.}
    \label{fig:agregate_bandwidth_results}
\end{figure*}

Bandwidth, assessed via linear frequency sweeps, followed an increasing-then-decreasing trend with torque (Figure \ref{fig:agregate_bandwidth_results}). The original motor peaked at 1 Nm before declining, while the SEA Module in open-loop control showed a less pronounced trend. Maximum torque varied until power limits were reached. Summary statistics are in Table \ref{tab:bandwidth_summary_statistics}.

\begin{table}[h!]
  \centering
  \caption{Summary Bandwidth statistics for the four tested systems.}
  \label{tab:bandwidth_summary_statistics}
  \vspace{0.3in}
  \begin{tabular}{c c c c c}
    System & \(B_{avg}\) & \(B_{min}\) & \(B_{max}\) & \(T\left(B_{max}\right)\)\\
     & [Hz] & [Hz] & [Hz] & [Nm] \\
    \hline
    Motor & $5.122$ &$ 0.420$ & $10.32$ & $1.00$ \\
    Open Loop SEA & $4.780$ & $1.175$ & $8.125$ & $5.00$ \\
    Closed Loop SEA & $15.86$ & $1.100$ & $30.32$ & $1.25$ \\
    Closed Loop ATI & $22.43$ & $15.60$ & $28.17$ & $1.50$ \\
  \end{tabular}
\end{table}

In open-loop control, the SE element reduced average bandwidth from $5.12$ Hz to $4.78$ Hz and maximum from $10.32$ Hz to $8.12$ Hz, decreases of $6.64$\% and $21.3$\% respectively. Closed-loop force control increased bandwidth 231\% to $15.86$ Hz, while ATI Gamma feedback improved it 369\% to $22.43$ Hz. The highest recorded bandwidth was $30.32$ Hz with SEA closed-loop control, surpassing the ATI Gamma’s $28.17$ Hz by 7.63\%.

\section{Discussion} \label{Discussion}

\subsection{Bandwidth Relationship with Torque} \label{Bandwidth Relationship with Torque}

\begin{figure}[h!]
    \centering
    \includegraphics[width=\columnwidth]{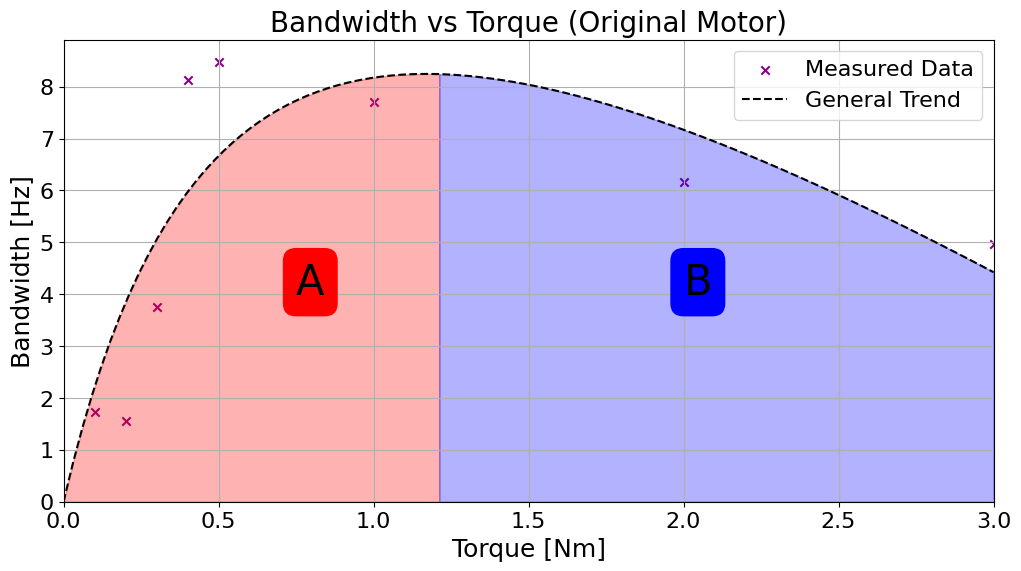}
    \caption{Original motor bandwidth-vs-torque measured data and trend line split into A (low-force bandwidth) and B (high-force bandwidth).}
    \label{fig:bandwidth_torque_regimes}
\end{figure}

The relationship between bandwidth and torque can be analyzed in two regimes, as shown in Figure \ref{fig:bandwidth_torque_regimes}. 

In the high-force regime (Section B), velocity saturation, rather than the system’s natural frequency, dominates bandwidth decay. As velocity increases, maximum achievable force decreases until reaching saturation, beyond which no additional force is generated. Without extra force to counteract losses (e.g., friction), velocity ceases to increase. Robinson et al. \cite{Robinson1999SeriesRobot} provide a model for this in Equation \ref{eq:21}. Rearranging and setting \(f\) equal to the bandwidth at saturation, Equation \ref{eq:23} shows that bandwidth is inversely proportional to torque amplitude, explaining the observed decay.

\begin{equation} 
    \label{eq:21}
    \left|F_m\right|=
    \begin{cases}
    F_{sat}\left(1-\frac{V_m}{V_{sat}}\right)& \text{if } \left|V_m\right| \geq V_{sat}\\
    0,              & \text{else}
    \end{cases}
\end{equation}

\begin{equation} 
    \label{eq:23}
    f \propto \frac{V_{sat}}{T_d}
\end{equation}

In the low-force regime (Section A), bandwidth increases with torque due to static friction effects \cite{Robinson1999SeriesRobot}. Below a threshold, motion—and thus bandwidth—remains zero. Once static friction is overcome, torque accelerates the rotor, reducing friction’s relative impact. The required angular velocity peaks at zero crossings, Equation \ref{eq:22}, making bandwidth proportional to torque, Equation \ref{eq:24}.

\begin{equation} 
    \label{eq:22}
    \omega_{max} = \frac{2 \pi f T_d}{k_\theta}
\end{equation}

\begin{equation} 
    \label{eq:24}
    \dot{\omega}_{max} = \frac{4\pi^2f^2T_d}{k_{theta}}
\end{equation}

\subsection{Open Loop Bandwidth Reduction} \label{Open Loop Bandwidth Reduction}

Pratt and Krupp \cite{Pratt2004LateActuators} highlight that SEA reduces open-loop gain, allowing higher control gains without compromising stability, mitigating stiction effects. This explains why SEA exhibits lower open-loop bandwidth than the original motor, as confirmed by measured data.

\subsection{Closed Loop Bandwidth Increase}

Experimental results confirm that SEA feedback maintains and even enhances actuator bandwidth, increasing the average from $5.12$ Hz to $15.86$ Hz and maximum from $10.32$ Hz to $30.32$ Hz, a threefold improvement. This effectively transformed the actuator into a system with faster response characteristics.

However, closed-loop bandwidth naturally improves, making these increases potentially misleading. Robinson et al. \cite{Robinson1999SeriesRobot} define a dimensionless constant, \(\Omega\), representing the ratio of controlled to motor natural frequency, which quantifies bandwidth gain for a given Proportional Derivative (PD) controller in Equation \ref{eq:25}. With unity proportional control, bandwidth should increase by \(\sqrt{2}\), not \(\sim3\), suggesting additional contributing factors.

\begin{equation} 
    \label{eq:25}
    \Omega = \frac{\omega_c}{\omega_n} = \sqrt{K_p+1}
\end{equation}

One explanation is that the linear model does not account for static friction or backlash. While static friction is low relative to torque amplitudes ($44$ mNm vs \(\sim1-5\) Nm), backlash is significant compared to SE element deflections. Baek et al. \cite{BaekAnalysisSystem} demonstrate that "backlash has a significant effect on the bandwidth of a system." Since the SEA measures instantaneous torque across the deforming element, it remains unaffected by backlash. The original motor may achieve bandwidths near $20$ Hz (\(\sqrt{2}\) lower than feedback results), but stiction significantly reduces open-loop bandwidth.

Higher bandwidths could theoretically be achieved by increasing the proportional control constant, which was attempted. However, stable gains were limited to \(\sim1.4\). Thus, this study focused on comparative analysis rather than maximizing absolute bandwidth.

To validate these gains, SEA feedback was replaced with ATI Gamma torque feedback. The SEA module (including PCB, bearings, and fasteners) cost \(\sim\)£25, while the ATI Gamma costs \$5260–\$7010 \cite{ATI:Series}. Despite expectations that the ATI Gamma would outperform the SEA, the SEA module achieved a maximum bandwidth of $30.32$ Hz, compared to $28.17$ Hz with the ATI Gamma—a $7.63\%$ improvement at just $0.35\%$ of the cost. The ATI Gamma’s rigidity amplifies backlash-induced impacts at motor dead zones, whereas the SEA module, measuring torque directly across the deforming element, is less affected.

However, the ATI Gamma provided greater consistency, with an average bandwidth of \(22.42\) Hz, a \(41.4\%\) improvement over the SEA module's \(15.86\) Hz. Additionally, its narrower bandwidth range (\(B_{min} = 15.60\) Hz to \(B_{max} = 28.17\) Hz) contrasts with the SEA module's broader variation (\(B_{min} = 1.10\) Hz to \(B_{max} = 30.32\) Hz), highlighting the ATI Gamma’s superior precision and signal processing.

These results underscore the trade-off between cost and performance. The SEA module offers a cost-effective alternative with competitive peak bandwidth but reduced consistency compared to industrial-grade sensors. Despite its lower cost, it exceeded the ATI Gamma's maximum bandwidth by \(7.63\%\) but exhibited greater variability.

\section{Conclusion} \label{Conclusion}

Herewith we demonstrated the implementation and bandwidth performance of an SEA retrofitted to a black-box motor. While SEAs typically reduce system bandwidth, this study confirmed that where non-linearities are present, it can be maintained or increased. The SEA was designed for force-control bandwidth, using an FSEA configuration to ensure torque measurements reflected the output load. A laser-cut torsional disc provided high precision, repeatability, and a well-defined mechanical response.

A proof-of-concept SEA module was retrofitted to an RMD X8 V2 actuator. The SE element, informed by FEA, exhibited a stiffness of \(2155.4\) Nm/rad, deviating \(-5.65\%\) from FE predictions. Displacement was measured with an AS5311 magnetic encoder at \(1.109\times10^{-5}\) rad/pulse resolution and calibrated to minimize uncertainties.

Bandwidth was determined via a linear sine sweep. With SEA feedback, maximum bandwidth increased from \(10.32\) Hz to \(30.32\) Hz, surpassing the ATI Gamma’s \(28.17\) Hz. These results support the hypothesis that an SEA can achieve high force-control bandwidth and even outperform rigid force-torque transducers by mitigating system non-linearities like backlash.

Future work includes advanced control strategies like observer-based designs and improving SEA manufacturing to reduce non-linearities and enhance consistency.

\bibliographystyle{IEEEtran}

\bibliography{references.bib}

\newpage

\section{Biography Section}

\begin{IEEEbiography}[{\includegraphics[width=1in,height=1.25in,clip,keepaspectratio]{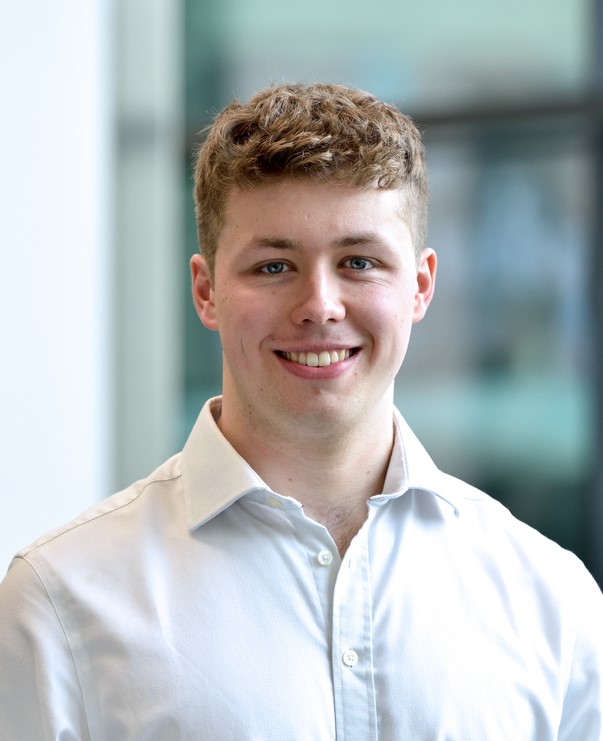}}]{Ivan Tregear}
Ivan Tregear (Member, iMeche) received the M.Eng. degree in mechanical engineering from Imperial College London, London, U.K., in 2023. 

He is currently the Chief Technology Officer of KAIKAKU.AI, a robotics company focused on revolutionizing the food and beverage industry by creating world-class restaurant experiences through the integration of hardware, software, and artificial intelligence.
\end{IEEEbiography}

\begin{IEEEbiography}[{\includegraphics[width=1in,height=1.25in,clip,keepaspectratio]{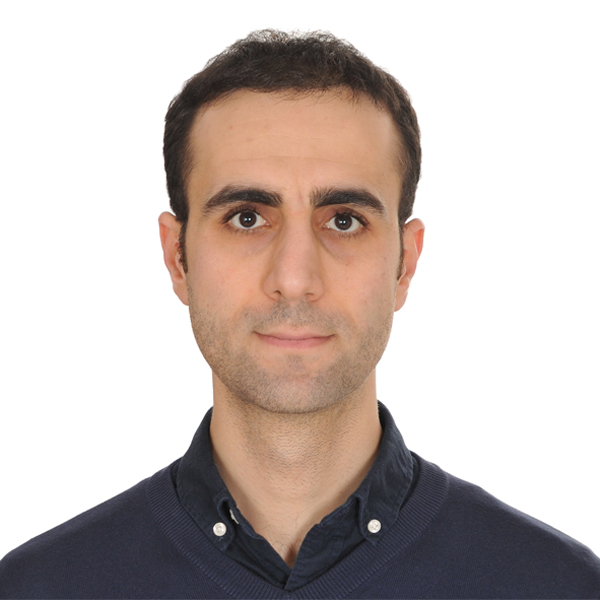}}]{Ayhan Aktas} received his M.Sc. degree in System Dynamics and Control from Yildiz Technical University, Istanbul, Türkiye, in 2018, and his Ph.D. degree in Surgical Robotics from Imperial College London in 2025.  He is currently a Research Associate at the Material Robotics Laboratory, Boston University.  His research interests include robotics, control systems, mechanical design, and soft robotics, with a particular focus on the application of robotic technologies in medical and surgical contexts.
\end{IEEEbiography}

\begin{IEEEbiography}[{\includegraphics[width=1in,height=1.25in,clip,keepaspectratio]{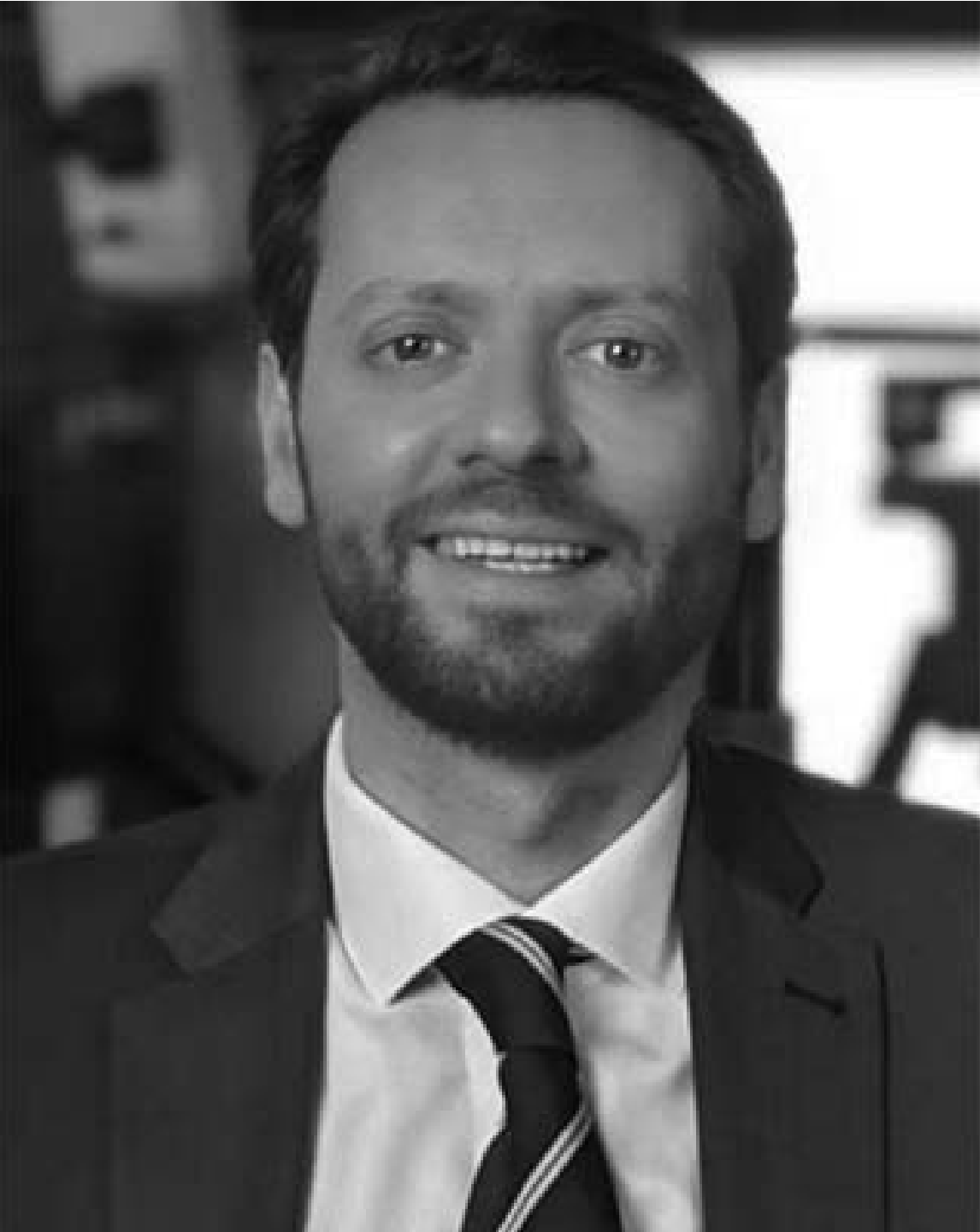}}]{Ferdinando Rodriguez y Baena}
(Member, IEEE) received the M.Eng. degree in mechatronics and manufacturing systems engineering from King’s College London, London, U.K., in 2000, and the Ph.D. degree in medical robotics from Imperial College London, London, in 2004.

He is currently a Professor in medical robotics with the Department of Mechanical Engineering, Imperial College London, where he leads the Mechatronics in Medicine Laboratory. He is also the Co-Director of the Hamlyn Centre, Imperial College London. His research interests include the application of mechatronic systems to medicine, in the specific areas of clinical training, diagnostics, and surgical intervention.
\end{IEEEbiography}

\vfill
\end{document}